\pdfoutput=1

\documentclass[11pt]{article}

\usepackage[preprint]{acl}

\usepackage{times}
\usepackage{latexsym}
\usepackage{graphicx}
\usepackage{booktabs}
\usepackage{subcaption}
\usepackage{multirow}
\usepackage{adjustbox}
\usepackage{url}
\usepackage{enumitem}

\usepackage[T1]{fontenc}

\usepackage[utf8]{inputenc}

\usepackage{microtype}

\usepackage{inconsolata}

\title{\textsc{NLAS-multi}: A Multilingual Corpus of Automatically Generated Natural Language Argumentation Schemes}

\author{Ramon Ruiz-Dolz$^{\dagger}$, Joaquin Taverner$^{\ddagger}$, John Lawrence$^{\dagger}$, \and Chris Reed$^{\dagger}$ \\
         $^{\dagger}$Centre for Argument Technology, University of Dundee, Dundee DD1 4HN, United Kingdom \\ 
         \{rruizdolz001, j.lawrence, c.a.reed\}@dundee.ac.uk \\
         $^{\ddagger}$VRAIN, Universitat Politècnica de València, 46022, València, Spain \\
         joataap@dsic.upv.es \\
         }

\begin{document}
\maketitle
\begin{abstract}
Some of the major limitations identified in the areas of argument mining, argument generation, and natural language argument analysis are related to the complexity of annotating argumentatively rich data, the limited size of these corpora, and the constraints that represent the different languages and domains in which these data is annotated. To address these limitations, in this paper we present the following contributions: (i) an effective methodology for the automatic generation of natural language arguments in different topics and languages, (ii) the largest publicly available corpus of natural language argumentation schemes, and (iii) a set of solid baselines and fine-tuned models for the automatic identification of argumentation schemes.
\end{abstract}

\section{Introduction}

The automatic generation of natural language arguments is a research topic that has gained a lot of popularity in the recent years. Specifically since the emergence of neural network-based models in the Natural Language Processing (NLP) domain. One of the first approaches using a neural architecture for natural language argument generation was proposed in \cite{le2018dave}, where a Long Short-Term Memory (LSTM) encoder-decoder architecture was trained for the argument generation task in a dialogue-like setup. This architecture was also used in \cite{hua2019argument} to generate arguments combining information retrieval, planning, and realisation, and in \cite{hua2018neural} where the LSTM was complemented with external evidence to generate new arguments. A different approach was proposed in \cite{wachsmuth2018argumentation}, where the authors propose a method based on the three Aristotelian rhetorical strategies of ethos, pathos, and logos. Similar approaches used an encoder-decorder architecture but based on the gated recurrent unit instead of the LSTM to address the automatic generation of arguments \cite{hidey2019fixed} reducing the amount of training parameters. In addition to these research trends, some other works explored the automatic generation of arguments with the use of pre-defined adaptable templates \cite{thomas2018argumessage}, with the inclusion of knowledge graphs into the proposed algorithms \cite{al2021employing}, and also with the use of generative language models such as GPT-2 \cite{gretz2020workweek}, the predecessor of the models researched in this paper. An original approach was proposed in \cite{el2019computational} where the synthesis of new arguments was done as it is done in standard language modelling. For that purpose, argumentative discourse units were treated as words and arguments as sentences. With the proposed argument language model it was possible to produce new arguments in a similar way as the language models generate text. In recent research the focus has been on new methods for the generation of counter-arguments \cite{jo2021knowledge, alshomary2021counter}, as it limits and specifies the process of generation and allows to improve the quality and usefulness of the generated arguments. More versatile approaches have also been explored, for example in \cite{schiller2021aspect, saha2023argu} where controllable frameworks are proposed in which parameters of the argument such as the topic, the stance, or the aspect can be indicated as a control code interpreted by the model. However, most of these automatic generation approaches are constrained to specific domains and languages included in their fine-tuning, and its difficult to use them in new domains and/or languages \cite{toledo2020multilingual}.

Furthermore, argument-based natural language corpora have been mostly annotated in English considering only the premise/claim classes of argument and attack/support types of argumentative relations \cite{stab2014annotating, peldszus2015annotated}. This annotation is interesting from the general argumentative discourse viewpoint, since it supports drawing an informative analysis of the basic elements and the overall structure of argumentation, but at the same time it represents an important limitation in terms of the level of detail at which the analysis can be carried out. In recent work, an effort has been made to further deepen the analysis of argumentative discourse in dialogues \cite{ruiz2021vivesdebate, hautli2022qt30}, but they present similar limitations with respect to the level of detail of the annotated arguments. The lack of a detailed annotation framework based on a solid model of argumentation makes it difficult to automate the analysis of more complex argumentative aspects such as the validity of the argumentative reasoning and its natural language content. In this respect the argumentation scheme model proposed by D. Walton \citet{walton2008argumentation} provides a good balance  between formality and natural language, and represents an opportunity to increase the degree of detail of natural language annotations in the argumentation domain. However, only a few corpora containing annotations based on the argumentation scheme model exist \cite{feng2011classifying, green2015identifying, musi2016towards, lawrence2019online, visser2020annotating, jo2021classifying}. Furthermore, given the high complexity of doing human annotation of argumentation schemes, their size and variety (i.e., topic, domain, stance) tends to be rather limited. Only a recent paper in the intersection of natural language argument generation and argumentation schemes claims to publicly share the largest argumentation scheme corpus \cite{saha2023argu} consisting of almost 70,000 arguments. The arguments included in this corpus are, however, short sentences labelled with an argumentation scheme-related label, without any formal structure (as argumentation schemes do) and labelling is not consistent with the argumentation scheme model. Therefore, the limitations related to the limited availability of in-depth annotated argumentative data remain unresolved. 

In this paper, we present three main contributions to overcome these limitations: (i) an effective argument structure-based methodology for the automatic generation of natural language arguments in different domains and languages without the need of fine-tuning, (ii) the largest publicly available corpus of natural language argumentation schemes (based on Walton's model), and (iii) a set of solid baselines and fine-tuned models that can serve as a reference for further research in more advanced argument mining tasks, such as the identification of argumentation schemes.

\section{Background}\label{sec:background}

The research work carried out in this paper has been theoretically grounded on two main concepts: (i) Large Generative Language Models and their subsequent prompting techniques \cite{zhao2023survey, liu2023pre}, and (ii) Walton's argumentation schemes \cite{walton2008argumentation}.

\subsection{Large Generative Language Models}

Since the proposal of the Transformer architecture \cite{vaswani2017attention}, research in natural language generation has been rapidly evolving and improving. State-of-the-art generative models such as GPT-4 \cite{openai2023gpt4}, Llama2 \cite{touvron2023llama}, and BLOOM \cite{scao2022bloom} are Transformer-based causal decoders that have been pre-trained in predicting the next token in a sequence/document (i.e., causal language modelling), and fine-tuned in an additional task to improve their performance \cite{zhao2023survey}. A task that has been found to be very effective for fine-tuning these models after pre-training is the Reinforcement Learning with Human Feedback (RLHF) \cite{christiano2017deep} which is used in the GPT-4 and Llama2 models. The main purpose of the RLHF task is to improve the \textit{alignment} of the model with desired human preferences. For that purpose, the reward model is trained to select the best output in a dialogue-like setup. This best output is defined based on previously collected human feedback. The resulting model is therefore capable of not only generating the next sequence of tokens in a document, but also choosing the most suitable option for a given situation in a dialogue-like setup. This is what we know as \textit{prompting} \cite{liu2023pre}, where the model is provided with a document or a sequence of tokens (i.e., the \textit{prompt}), and outputs a (natural language generated) response to it. Given the generalised promising results observed in \textit{prompt}-based approaches for different NLP tasks (e.g., text classification \cite{zhang2022prompt}, machine translation \cite{li2022prompt}, or event extraction \cite{si2022generating} among others), we propose the adoption of a \textit{prompt}-based paradigm combined with the structured argumentation model defined by the argumentation schemes as the next step in the area of natural language argument generation.

\subsection{Walton's argumentation schemes}

The argumentation schemes proposed and defined by Walton provide a mature and well-researched model of argument representation in which the natural language and the logical structure of the argument can be partially dissociated. The argumentation scheme model can be understood as as an intermediate representation between formal and informal representations of human argumentation. While research in formal argumentation (mostly) dispenses with natural language \cite{dung1995acceptability, baroni2011introduction}, informal argumentation research is mainly based in natural language instances of human argumentation \cite{bex2013implementing, hua2019argument, lawrence2020argument}. However, argumentation schemes combine aspects from formal logic and abstract representations with natural language to define different types of argumentative reasoning patterns. Let us consider the \textit{Argument from Position to Know} scheme as an example, Walton defines this argument as \cite{walton2008argumentation}:

\vspace{-0.5em}

\begin{itemize}[leftmargin=*]
\setlength\itemsep{-0.3em}
    \item[] \underline{\textit{Major Premise}}: Source $s$ is in position to know about things in a certain subject domain $f$ containing proposition $p$.
    \item[] \underline{\textit{Minor Premise}}: $s$ asserts that $p$ is true (false).
    \item[] \underline{\textit{Conclusion}}: $p$ is true (false).
\end{itemize}

\vspace{-0.5em}




An argumentation scheme thus provides a set of abstract variables 
(i.e., $s$, $f$, and $p$ in this case) that can be replaced with natural language text, together with the connections (in natural language) between these variables required to make it each specific argumentation scheme. This makes the argumentation scheme model perfect to combine it with the state-of-the-art natural language generation models to approach the automatic generation of natural language arguments from a new paradigm that we propose in this paper. More than sixty types of argumentation schemes representing the most common patterns of human reasoning were compiled in \cite{walton2008argumentation}, from which we selected the twenty most commonly used in human argumentation for our experiments as described in the following section.

\section{Automatic Generation of NLAS}





In this section, we describe the process used to build the corpus through the generation of Natural Language Argumentation Scheme (NLAS) using the APIs of \textsc{GPT-3.5-turbo} and \textsc{GPT-4}. To do so, we selected 20 different argument schemes, included in Appendix \ref{app:schemes} together with their respective acronyms. Furthermore, each argument was instantiated with 50 different topics including claims referring to euthanasia, mandatory vaccination in pandemic, climate change, and UFO existence (see Appendix \ref{app:topics} for the complete list of topics). Finally, each argument was generated with two stances: in favour and against. This resulted in the generation of 2000 arguments (20 schemes x 50 topics x 2 stances) for two languages: English and Spanish.

\begin{figure}[htb]
    \centering
    \includegraphics[width=\columnwidth]{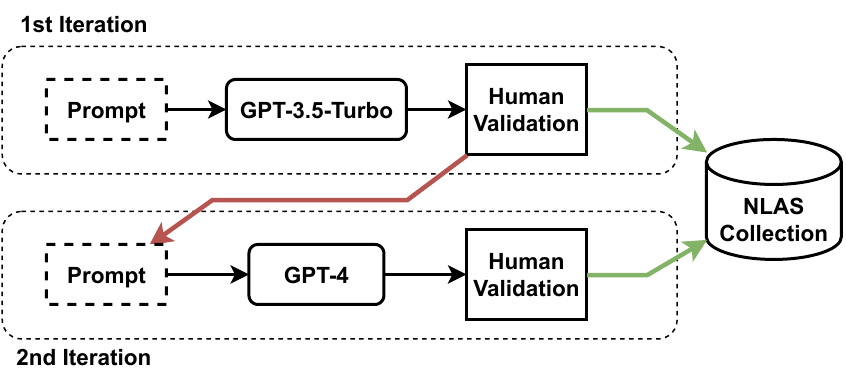}
    \caption{Automatic Generation of NLAS}
    \label{fig:method}
\end{figure} 

Figure~\ref{fig:method} illustrates the procedure followed to generate the \textsc{NLAS-multi} corpus. This procedure is divided into two iterations. In a first iteration, we generated 2000 arguments for each language (English and Spanish) using \textsc{GPT-3.5-turbo}. These arguments were evaluated by five expert annotators: three for English and two for Spanish. During this evaluation, the experts considered only the adequacy of the argument generated in relation to the pattern of each argumentation scheme, the topic, and the stance and annotate these arguments as valid or not valid. Arguments that were classified as not valid (i.e., arguments that did not adhere to the argumentative structure, topic, or stance) were discarded. 

In a second iteration, \textsc{GPT-4} was used to re-generate the arguments that were discarded in the first iteration using the same prompt design. Expert annotators evaluated the generated arguments using the same criteria as in the evaluation of the first iteration. Finally, arguments that were classified as suitable by the expert evaluators in both iterations were used to generate the \textsc{NLAS-multi} corpus. Note that, our objective is to obtain a labelled corpus, i.e., we are not evaluating the GPT-3.5-turbo and GPT-4 accuracy. Therefore, this two-step methodology is justified by the reduction of the economic, temporal and computational cost.  

\subsection{Natural Language Generation}







To generate the arguments, we followed a communication strategy with the APIs of \textsc{GPT-3.5-turbo} and \textsc{GPT-4}, applying prompting techniques.
With these techniques, we aimed to maintain the original structure of each argument scheme proposed by Walton. To achieve this, the general structure of our prompt comprised three main parts: the argument type specification, the argument pattern, and the output format.

For the argument type specification, we request that the system generates a specific argument type (e.g., Argument from Position to Know) by defining the stance (i.e., in favour or against) and the topic (e.g., euthanasia). Following these guidelines, an example of the argument type specification would be: `Provide a position to know argument in favour of euthanasia'. Next, we attached the template of the argument scheme pattern to be instantiated. The scheme was presented in accordance with the specifications proposed by Walton for each type of argument (see the example of the specification of the Argument from Position to Know in Section~\ref{sec:background}). 

Finally, in the prompt, we specified the desired output format to be provided in response to the request. We requested that the output should be in JSON format, maintaining the structure of the argumentation schema provided.

The design of the prompt is based on the work proposed in \cite{white2023prompt,liu2022design}. We conducted an initial round of exploratory experiments to test the design of the communication with the API through prompts. The objective was to establish prompts to generate arguments on different topics and stances, while complying with the structure of the argumentation schemes. However, considering some issues in the formal logic consistency of argumentation schemes identified in previous research \cite{verheij2003dialectical}, this initial round of experiments produced unsatisfactory outcomes. In some argumentation schemes, the system was unable to identify the variables that needed to be instantiated. Taking the example of Argument from Position to Know, the language model often fails to recognise the variable $s$ as a candidate for being instantiated, resulting in the return of the argument structure without instantiating that variable. To address this problem, we adopted the approach of using the structure `[variable]' to denote the variables as discussed in \cite{jiang2022promptmaker}. In an initial attempt, we introduced square brackets to the variables within Walton's original scheme (e.g., `source $s$' was revised to `source [$s$]'). While this adjustment significantly enhanced the system's capability to instantiate arguments, issues still persisted when instantiating the some variables. These problems arose due to the system's inability to discern which element to instantiate within each variable. Therefore, we opted to redefine the specification of the argument schemes to make the variables more explicit while maintaining the consistence of the argumentation scheme. Thus, the scheme for the Argument from Position to Know was described as:

\vspace{-0.5em}

\begin{itemize}[leftmargin=*]
\setlength\itemsep{-0.3em}
    \item[] \underline{\textit{Major Premise}}: \textbf{[Someone]} is in position to know about things in a certain subject domain \textbf{[Domain]} containing proposition \textbf{[A]}.
    \item[] \underline{\textit{Minor Premise}}: \textbf{[Someone]} asserts that \textbf{[A]} is true.
    \item[] \underline{\textit{Conclusion}}: \textbf{[A]} is true.
\end{itemize}

\vspace{-0.5em}

This way, the system was able to identify that the variable [Someone] in the \textit{Major Premise} should be modified by the name of a person (or the name of a group or collective) and use that name in the \textit{Minor Premise}.

Once the structure of the prompt was determined, we generated the 2,000 arguments for English and Spanish using the APIs of \textsc{GPT-3.5-turbo} and \textsc{GPT-4} according to the process described above. 


The following is an example of the argument generated for the Argument From Position to Know scheme on the topic of euthanasia and with the stance in favour using \textsc{GPT-3.5-turbo}:

\vspace{-0.5em}

\begin{itemize}[leftmargin=*]
\setlength\itemsep{-0.3em}
    \item[] \underline{\textit{Major Premise}}: Medical professionals (such as doctors and nurses) are in position to know about the treatment options available for terminally ill patients.
    \item[] \underline{\textit{Minor Premise}}: Many medical professionals argue that euthanasia is a humane option for terminally ill patients who are experiencing unbearable suffering and have little hope for recovery.
    \item[] \underline{\textit{Conclusion}}: Euthanasia can be a morally justifiable option for terminally ill patients who are experiencing unbearable suffering and have little hope for recovery.
\end{itemize}

\vspace{-0.5em}

As can be seen in this example, \textsc{GPT-3.5-turbo} was able to modify the syntactic structure of the argument, enriching the content, without affecting the semantics or the structure of the argument scheme. 

\subsection{Human Validation}

After the natural language generation of arguments, two human validation steps were carried out. For that purpose, we hired five linguists that were familiar with the theory of argumentation schemes and we provided them with the formal definitions of the 20 argumentation schemes included in our experiments. Their task was to check whether a NLAS was valid or not considering its argumentation scheme structure, its topic, and its stance. Only in the case of an argument following the expected argumentation scheme structure, arguing about the expected topic, and aligned with the expected stance was considered as valid. If any of these variables was violated, the argument was annotated as non-valid and rejected from our collection. It is important to emphasise that aspects such as the soundness, persuasiveness, or the strength of an argument were not considered in our annotation. Therefore, some arguments that might not be perceived as solid by humans were still labelled as valid as long as they respected all the variables described as relevant. Four of the five annotators worked on the English arguments and two of them worked on the Spanish ones. One bilingual annotator worked in both English and Spanish. 

\begin{figure*}
\centering
     \begin{subfigure}{0.4\textwidth}
         \includegraphics[width=\linewidth]{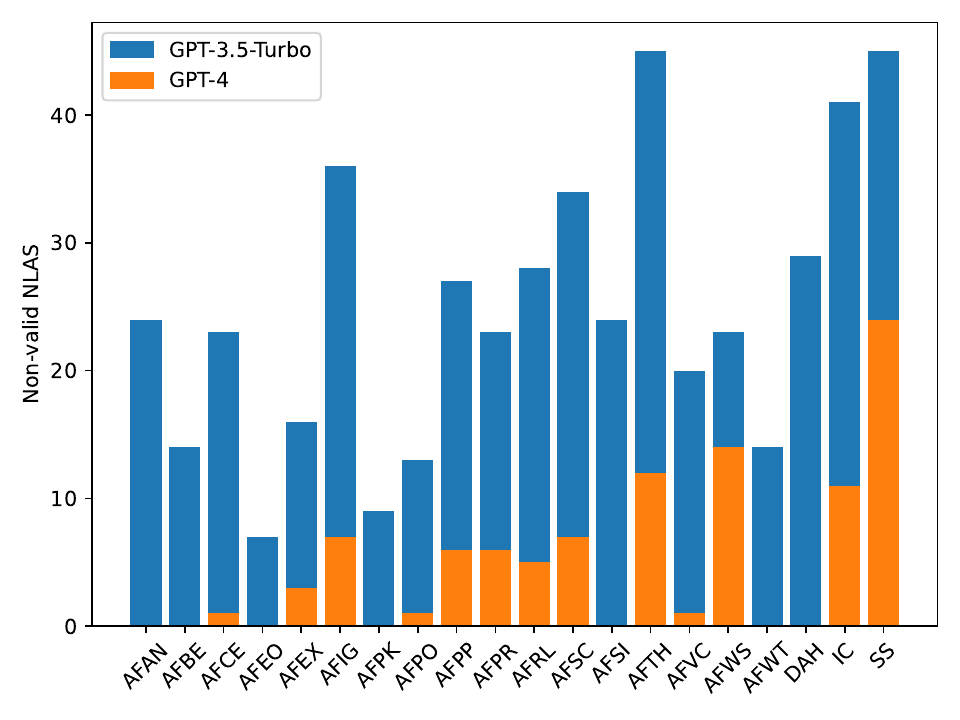}
         \caption{Non-valid English NLAS}
         \label{fig:eng_err}
     \end{subfigure}
     \begin{subfigure}{0.4\textwidth}
         \includegraphics[width=\linewidth]{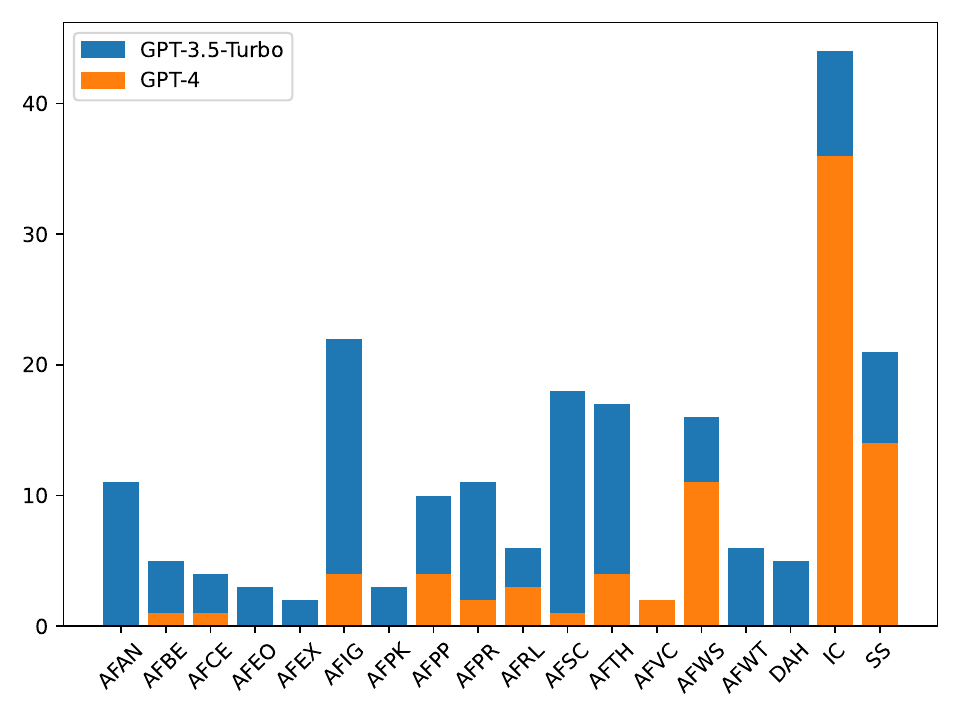}
         \caption{Non-valid Spanish NLAS}
         \label{fig:esp_err}
     \end{subfigure}
        \caption{Error distribution of the generated NLAS (lower is better).}
        \label{fig:err_distrib}
\end{figure*}

In the first human validation step, we analysed the NLAS produced in the first natural language generation iteration with \textsc{GPT-3.5-turbo}. Therefore, in this first step, 2,000 natural language arguments were manually annotated in English, and another 2,000 in Spanish. Regarding the English part we obtained 
1,496 valid and 
504 non-valid NLAS, meaning that with \textsc{GPT-3.5-turbo} we were able to achieve an initial accuracy of 74.8\%. From this first validation, we observed that the error distribution was not uniform, some argumentation schemes were more challenging than others during the natural language generation part of the process (see Figure \ref{fig:eng_err}, blue bars). 
Regarding the Spanish part, we obtained 1,794 valid and 206 non-valid NLAS, achieving an initial accuracy of an 89.7\%. In the Spanish part, we observed a significant smaller amount of non-valid NLAS than in English, and the error distribution was mainly dominated the argument from inconsistent commitment (see Figure \ref{fig:esp_err}, blue bars). The error distribution of the other argumentation schemes was significantly lower than in English. 



In the second human validation step, we analysed the NLAS produced in the second natural language generation iteration with \textsc{GPT-4}. Compared to the first step, in this second step we only re-generated the arguments that were annotated as non-valid in the first iteration. Therefore, 
504 NLAS were manually annotated in English and 206 in Spanish in the second human validation step. As for the English part, in the end of this second step we obtained 397 more valid NLAS and 
107 non-valid arguments. Compared to the first iteration, \textsc{GPT-4} was able to improve the performance by generating correctly a 
78.8\% of the natural language arguments that \textsc{GPT-3.5-turbo} failed to generate as intended. This was a big improvement, allowing us to correctly generate a total of 1,893 NLAS out of the 2,000 prompted arguments (i.e., a 
94.7\%) following our proposed methodology. The errors identified in this second validation step, however, were also not distributed uniformly between the different types of argumentation schemes (see Figure \ref{fig:eng_err}, orange bars). 
As for the the second iteration with the Spanish part, we obtained 123 valid and 83 non-valid NLAS, achieving an accuracy of 60\% with \textsc{GPT-4}. This improvement allowed us to generate a total of 1,917 valid NLAS, a 95.8\% of the prompted argumentation schemes. Again, the error in this second validation step was focused on the inconsistent commitment argumentation scheme (see Figure \ref{fig:esp_err}, orange bars). It is worth to mention that the conclusion of the inconsistent commitment scheme is that somebody's commitments are inconsistent. This scheme has two commitment premises, which conflict with each other. Therefore, it is hard to define the stance of an inconsistent commitment NLAS. In order to avoid having redundant information and repeated arguments in our collection, we defined the validity of an inconsistent commitment NLAS if the first commitment premise is aligned with the stance of our prompt. Therefore, even though we rejected a large number of inconsistent commitment arguments compared to others, this happened mostly due to an effort to minimise repeated NLAS in our collection than due to errors in the logical structures of the automatically generated arguments.




To validate our annotation, we carried out two Inter Annotator Agreement (IAA) studies containing a 10\% subset of the total number of samples, one for each language (i.e., 200 NLAS per language). Regarding the IAA in English, we report a 0.65 Cohen's Kappa \cite{cohen1960coefficient}, meaning that there was a substantial agreement between the four annotators. On the other hand, we report a 0.22 Cohen's Kappa in the Spanish annotation, meaning that our annotators reached only a fair agreement. The agreement in the Spanish part of the corpus was significantly lower than the one observed in the English part, however, we observed that the total number of disagreements was lower than in the English study (i.e., 21 disagreements in Spanish versus 23 in English). This variation can be attributed to the significantly low number of non-valid arguments annotated in the Spanish part compared to the English part of the corpus. A disagreement in a non-valid argument represents a higher ratio to the absolute number of invalid arguments compared to the English part, and this has a direct effect to the Cohen's Kappa score.

\subsection{The \textsc{NLAS-multi} Corpus}

\begin{table*}[]
\centering
\adjustbox{width=\textwidth}{%

\begin{tabular}{lrrrrlrrrr}
\toprule
 & \multicolumn{4}{c}{\textbf{English}} &  & \multicolumn{4}{c}{\textbf{Spanish}} \\ \cline{2-5} \cline{7-10} 
 & \multicolumn{1}{l}{\multirow{2}{*}{\textbf{Inferences}}} & \multicolumn{1}{l}{\multirow{2}{*}{\textbf{Words}}} & \multicolumn{2}{c}{\textbf{Arguments}} &  & \multicolumn{1}{l}{\multirow{2}{*}{\textbf{Inferences}}} & \multicolumn{1}{l}{\multirow{2}{*}{\textbf{Words}}} & \multicolumn{2}{c}{\textbf{Arguments}} \\ \cline{4-5} \cline{9-10} 
 & \multicolumn{1}{l}{} & \multicolumn{1}{l}{} & \multicolumn{1}{l}{\textbf{In favour}} & \multicolumn{1}{l}{\textbf{Against}} &  & \multicolumn{1}{l}{} & \multicolumn{1}{l}{} & \multicolumn{1}{l}{\textbf{In favour}} & \multicolumn{1}{l}{\textbf{Against}} \\\midrule \midrule
Total & 3,949 & 118,493 & 941 & 952 & \multicolumn{1}{r}{} & 4,015 & 135,023 & 974 & 943 \\ \midrule
Mean & 197.5 & 5,924.7 & 47.1 & 47.6 & \multicolumn{1}{r}{} & 200.8 & 6,751.2 & 48.7 & 47.2 \\ \midrule
Sd & 49.4 & 1,787.1 & 6.5 & 3.4 & \multicolumn{1}{r}{} & 54.5 & 1,993.4 & 3.2 & 7.1 \\ \bottomrule
\end{tabular}}
\caption{Description of the number of inferences, words, and arguments in the corpus for the 20 argumentation schemes. }
\label{tab:argumen stats}
\end{table*}

As a result of the previous natural language argument generation experiments and the subsequent human validation, we compiled the \textsc{NLAS-multi} corpus\footnote{\url{https://doi.org/10.5281/zenodo.8364002}}, a multilingual corpus of 3,810 automatically generated NLAS.

On the one hand, Table~\ref{tab:argumen stats} presents a brief description of the inferences, words, and NLAS generated for the 20 argumentative schemes in the \textsc{NLAS-multi} corpus. To understand the distribution of the \textsc{NLAS-multi} corpus, it is essential to emphasise that it has been generated from intentionally developed atomic arguments, with conflicting positions (i.e., arguments for and against for each argumentative scheme and each topic), and adapted to different argumentative schemes that include predefined premises and conclusions (refer to the argumentative scheme presented in Section~\ref{sec:background}). This directly affects the number of inferences, NLAS per topic, and conflicts, as it will be discussed below.

The \textsc{NLAS-multi} corpus consists of a total of 253,516 words distributed in 118,493 words in English and 135,023 words in Spanish. This makes our corpus one of the most extensive currently available. Comparing it with some of the largest publicly available corpora in the literature, we find that our corpus is significantly larger than the VivesDebate corpus \cite{ruiz2021vivesdebate} with 139,756 words, and slightly smaller than the QT30 corpus \cite{hautli2022qt30} with 279,996 words. 

In terms of inferences, our corpus has a total of 7,964 (3,949 in English and 4,015 in Spanish). This places our corpus below ViveDebate with 12,653 and above QT30 with 5,205 inferences. However, it is important to note that both VivesDebate and QT30 were extracted from videos or radio programs with real and open debates. In contrast, as previously mentioned, the \textsc{NLAS-multi} corpus consists of atomic arguments modelled from argumentative schemes with a predetermined structure (premises and conclusions). Thus, the number of inferences in our corpus is limited by the number of premises defined in each argument scheme, unlike VivesDebate and QT30, which allowed a free format for argumentation. Therefore, we can find that arguments such as AFEX or DAH which, having only one premise, present a lower number of inferences (97 and 100 respectively in English and 100 in both cases in Spanish), while arguments such as AFBE or AFWT, which have three premises, have a higher number of inferences (300 both in English and, 297 and 300 in Spanish respectively). 

Finally, regarding the stance, in English, approximately 49.7\% of the arguments are in favour, while 50.3\% are against. In Spanish, the proportion is 50.8\% of arguments in favour and 49.2\% against. In both languages most of the arguments have around 50 samples of each type. 

\begin{table*}[]
\centering
\adjustbox{width=\textwidth}{%

\begin{tabular}{lrrrrlrrrr}
\toprule 
 & \multicolumn{4}{c}{\textbf{English}} & \textbf{} & \multicolumn{4}{c}{\textbf{Spanish}} \\ \cline{2-5} \cline{7-10} 
 & \multirow{2}{*}{\textbf{Inferences}} & \multirow{2}{*}{\textbf{Conflicts}} & \multicolumn{2}{c}{\textbf{Arguments}} & \textbf{} & \multirow{2}{*}{\textbf{Inferences}} & \multirow{2}{*}{\textbf{Conflicts}} & \multicolumn{2}{c}{\textbf{Arguments}} \\ \cline{4-5} \cline{9-10} 
 &  &  & \textbf{In favour} & \textbf{Against} & \textbf{} &  &  & \textbf{In favour} & \textbf{Against} \\ \midrule \midrule
Total & 3,949 & 11,626 & 941 & 952 &  & 4,015 & 12,155 & 974 & 943 \\ \midrule
Mean & 79.0 & 232.5 & 18.8 & 19.0 &  & 80.3 & 243.1 & 19.5 & 18.9 \\ \midrule
Sd & 3.2 & 16.2 & 1.0 & 0.9 &  & 3.0 & 17.6 & 0.7 & 1.0\\ \bottomrule
\end{tabular}}
\caption{Description of number of inferences, conflicts, and arguments for the 50 topics.}
\label{tab:topic stats}
\end{table*}

On the other hand, Table~\ref{tab:topic stats} presents a description of the 50 topics included in the corpus. The variance in both English and Spanish is relatively low, indicating that most of the topics have a number close to 80 inferences. In the case of English, the minimum is 73 inferences for the topic `Physical appearance for personal success', while the maximum is 84 for topics such as `Mandatory vaccination in pandemic' or `Renewable energy'. Similarly, in Spanish, the minimum is 74 inferences for topics such as `Freedom of speech' or `Climate change', while the maximum is 84 for topics such as `Internet censorship' or `Surrogacy'. 

Furthermore, the \textsc{NLAS-multi} corpus contains a total of 23,781 conflict relations. It is important to note that in order to calculate these conflicts, the argumentation schemes DAH, IC, AFPR, and AFTH were excluded, since their conclusions are not directly related to the acceptance or rejection of the topic.
Thanks to the strategy employed in the development of the corpus, which involves the use of conflicting positions for each argumentative schema and each topic, our corpus has a significantly higher number of conflicts compared to previous corpora in the literature. For example, the QT30 corpus contains only 976 conflicts, while the VivesDebate corpus is limited to 1,558.

Finally, the distribution of NLAS in favour and against the 50 topics is close in both cases to 19, with standard deviations of 1 or less (the maximum value of 20 corresponds to the 20 argumentation schemes). The distribution of NLAS is relatively uniform, with the minimum being 16 for the topic `Physical appearance for personal success' in English, and 15 for the topic `Climate change' in Spanish.



\section{Automatic Classification of NLAS}

To complement the automatic generation of natural language arguments, we provide a set of experimental results and baselines on the task of automatic classification of NLAS. Previous to the publication of our corpus, the automatic classification of argumentation schemes in natural language text has been an under-researched task considering only small sets of natural language instances and schemes \cite{feng2011classifying, lawrence2016argument}. This is mainly due to the difficulties of collecting and annotating natural language data containing argumentation schemes, caused by the elevated complexity for humans to annotate this information. However, with our NLAS corpus, we can now explore the capabilities of state-of-the-art algorithms when approaching this task in an instance with 20 different classes of argumentation schemes.

\subsection{Experimental Setup}

We approach the classification of NLAS as a sequence classification problem with a class dimensionality of 20 classes. Each argumentation scheme is represented as the concatenation of its argumentative components. For example, in the \textit{Argument from Position to Know} scheme we consider all the natural language text belonging to the \textit{Major Premise}, \textit{Minor Premise}, and \textit{Conclusion} as our natural language input to be classified into one of the 20 different classes of argumentation schemes included in the \textsc{NLAS-multi} corpus.

In all of our experimental results we have followed the same methodology. First, we used a RoBERTa \cite{liu2019roberta} architecture as our starting point. We have experimented with different versions of the model for the different experiments. We used the pre-trained (in English) RoBERTa-large \cite{liu2019roberta} model for the English experiments, the pre-trained (in Spanish) BERTIN-RoBERTa-base \cite{de2022bertin} model for the experiments in Spanish, and pre-trained (in more than 100 languages) XLM-RoBERTa-large \cite{conneau2020unsupervised} model for the bilingual experiments. We called these models\footnote{\url{https://huggingface.co/raruidol}} RoBERTa-eng, RoBERTa-esp, and RoBERTa-multi respectively. Thus, in our experiments we considered three different data configurations: (i) 1,893 NLAS in English only, (ii) 1,917 NLAS in Spanish only, and (iii) 3,810 NLAS in both languages. In all the three data configurations we divided our samples into train, development, and test following the 80-10-10 proportion respecting the class distribution in all the data splits.

We have used the same hyperparameter configuration in all of our experiments. We fine-tuned each model on its corresponding data configuration for 5 epochs for the RoBERTa-eng and RoBERTa-multi, and 10 epochs for the RoBERTa-esp with a batch size of 32, 24, and 128 respectively, a learning rate of 1e-5, and a weight decay of 0.01. All our models were trained using an Intel Core i7-9700k CPU with an NVIDIA RTX 3090 GPU and 32GB of RAM.

\subsection{Results}

\begin{table}
    \centering
    \resizebox{\columnwidth}{!}{%
    \begin{tabular}{l c c c}
    \toprule
    \textbf{Experiment} & \multicolumn{3}{c}{\textbf{Evaluation Metrics}} \\ \cline{2-4}
    \textbf{Model} &  \textbf{Precision} & \textbf{Recall} & \textbf{macro-F1}\\ \midrule\midrule
    RoBERTa-\textsc{eng} & 99.4 & 99.3 & 99.3 \\ \midrule 
    RoBERTa-\textsc{esp} & 97.6 & 97.4 & 97.4 \\ \midrule 
    RoBERTa-\textsc{multi} & 99.5 & 99.5 & 99.5 \\ \bottomrule
    \end{tabular}}
    \caption{Results of the Argumentation Scheme classification task. The reported results have been averaged from 3 randomly initialised sequential runs.}
    \label{tab:res1}
\end{table}

We have evaluated our models following two different strategies. In the first evaluation, we considered a homogeneous division of the data. This means that our samples were divided into train, development, and test without considering their topic or their stance (i.e., samples belonging to the same topic can be in any of the three splits). The results of this first evaluation have been summarised in Table \ref{tab:res1}. As we can observe the observed results were very good in all the linguistic setups. Only a small drop of performance can be observed in the Spanish experiments, but that is probably due to the fact that the model is a RoBERTa-base instead of a RoBERTa-large (as it is in the English and multilingual experiments).

\begin{table}
    \centering
    \resizebox{\columnwidth}{!}{%
    \begin{tabular}{l c c c}
    \toprule
    \textbf{Experiment} & \multicolumn{3}{c}{\textbf{Evaluation Metrics}} \\ \cline{2-4}
    \textbf{Model} &  \textbf{Precision} & \textbf{Recall} & \textbf{macro-F1}\\ \midrule\midrule
    RoBERTa-\textsc{eng} & 99.7 & 99.7 & 99.6 \\ \midrule 
    RoBERTa-\textsc{esp} & 97.4 & 96.2 & 97.0 \\ \midrule 
    RoBERTa-\textsc{multi} & 99.2 & 99.2 & 99.1 \\ \bottomrule
    \end{tabular}}
    \caption{Results of 5-fold topic-wise evaluation of the Argumentation Scheme classification task.}
    \label{tab:res2}
\end{table}

Given these strong results in our first evaluation, in the second evaluation strategy, we wanted to make the task more challenging for our models by considering a topic-wise division of the data. For that purpose we measured the performance of the models as the average of a 5-fold evaluation where each fold is defined on the basis of the set of topics. This way, in each iteration, all the NLAS belonging to 40 topics are included in the training split and the NLAS belonging to 5 new topics to both the development and test splits. Therefore, the topics included in development and test are not seen during training. The observed results in this second evaluation have been summarised in Table \ref{tab:res2}. As we can observe the results were almost as good as the previous ones (even better in the case of our English setup). Although we are addressing a 20 class classification NLP problem that has its roots in argumentation theory, natural language differences between an argumentation scheme and another are strong enough to make the task easier for a fine-tuned LLM such as RoBERTa. 


\section{Discussion}

In our paper, we present three main contributions to the computational argumentation community: (i) an original prompt-based methodology to effectively generate natural language argumentation schemes; (ii) the largest corpus of natural language argumentation schemes and one of the richest resources for natural language argumentative analysis composed of 3,810 arguments belonging to 20 different types of argumentative reasoning structures (i.e., argumentation schemes) instantiated in 50 different topics and 2 stances; and (iii) a set of solid baselines and outstanding results in the task of automatically classifying the reasoning structure of a natural language argument into the selected 20 classes of schemes.

With our experiments, we have demonstrated that effectively generating natural language arguments with a well defined structure is possible by prompting LLMs such as GPT-4. Our results, however, were far from being perfect with more than 5\% non-valid arguments after the two iterations in both languages. With our new human annotated corpus, we foresee as a future step in the direction of this paper to automatically supervise the automatic generation of natural language arguments with a model fine-tuned in our corpus together with a generative LLM. Furthermore, there is an additional and very important aspect that deserves discussion, it is the risks of applying our proposed method to generate arguments in sensitive contexts. With our method we do not have specific control of the natural language that is being produced and matched with the prompted structure. Therefore, in some contexts, the content of the produced arguments might be offensive, misleading or fallacious. 
A potential way of addressing this issue in future work will be to conduct a thorough analysis of the content of the NLAS (rather than just validating their structure), for which the critical questions of the argumentation schemes \cite{walton2008argumentation} will play a pivotal role.


\section*{Limitations}

The main limitation of this paper is related to the interpretation of the reported results in the atuomatic classification of NLAS. We would like to emphasise that the experimental results provided in the last section should be interpreted as a corpus-specific solid baseline. The reported outstanding results do not mean that our models can classify almost perfectly the 20 argumentation schemes in any context, but that \textbf{complete} natural language instances of the argumentation schemes can be effectively classified with our publicly available models. Humans, however, in order to ease and make our communication more fluid, make use of enthymemes\footnote{An enthymeme is an argument in which some element (e.g., a premise) has been omitted.}. The validation of our models in such challenging situations is out of the scope of this paper, but it could have a significant impact on their performance and it deserves a deeper analysis in future research.

\section*{Acknowledgements}

This work has been partially supported by the `AI for Citizen Intelligence Coaching against Disinformation (TITAN)' project, funded by the EU Horizon 2020 research and innovation programme under grant agreement 101070658, and by UK Research and innovation under the UK governments Horizon funding guarantee grant numbers 10040483 and 10055990. Also partially supported by Generalitat Valenciana CIPROM/2021/077, Spanish Government projects PID2020-113416RB-I00 and TED2021-131295B-C32; TAILOR project funded by EU Horizon 2020 under GA No 952215.

\bibliography{custom}

\appendix

\section{\textsc{NLAS-multi}: Argumentation Schemes}
\label{app:schemes}

We selected 20 different argument schemes, including: 

\begin{itemize}
    \item Argument From Analogy (AFAN)
    \item Argument From Best Explanation (AFBE)
    \item Argument From Cause to Effect (AFCE)
    \item Argument From Expert Opinion (AFEO)
    \item Argument From Example (AFEX)
    \item Argument From Ignorance (AFIG)
    \item Argument From Position to Know (AFPK)
    \item Argument From Popular Opinion (AFPO)
    \item Argument From Popular Practice (AFPP)
    \item Argument From Precedent (AFPR)
    \item Argument From Established Rule (AFRL)
    \item Argument From Sunk Costs (AFSC)
    \item Argument From Sign (AFSI)
    \item Argument From Threat (AFTH)
    \item Argument From Verbal Classification (AFVC)
    \item Argument From Waste (AFWS)
    \item Argument From Witness Testimony (AFWT)
    \item Direct ad Hominem (DAH)
    \item Inconsistent Commitment (IC)
    \item Slippery Slope (SS)
\end{itemize}

\section{\textsc{NLAS-multi}: Topics}
\label{app:topics}

Each argument was generated in English and Spanish languages with in favour and against stances for 50 different topics: 

\textit{Euthanasia}, \textit{Mandatory vaccination in pandemic}, \textit{Physical appearance for personal success}, \textit{Intermittent fasting}, \textit{Capital punishment}, \textit{Animal testing}, \textit{Climate change}, \textit{Legalisation of cannabis}, \textit{Abortion}, \textit{Freedom of speech}, \textit{Tax increase}, \textit{Animal/human cloning}, \textit{Research in artificial intelligence}, \textit{Nuclear energy}, \textit{Use of online social networks}, \textit{Gun control}, \textit{Universal basic pension}, \textit{Gender quotas}, \textit{Genetic manipulation}, \textit{Reduction in working time}, \textit{Remote work}, \textit{Increasing security by sacrificing individual privacy}, \textit{Censorship in social networks}, \textit{Terraplanism}, \textit{Renewable energy}, \textit{Electric transport}, \textit{Full self-driving cars}, \textit{Control measures to prevent economic inequality}, \textit{Immigration}, \textit{Offshore tax havens}, \textit{Tariffs on imported products}, \textit{Assisted suicide}, \textit{Birth control}, \textit{Globalisation}, \textit{Internet censorship}, \textit{Legalisation of prostitution}, \textit{Use of nuclear weapons}, \textit{Immortality}, \textit{Surrogacy}, \textit{Indiscriminate launching of satellites}, \textit{Drone strikes}, \textit{Internet access for children}, \textit{School uniform}, \textit{Regulation of unhealthy foods}, \textit{Political correctness}, \textit{UFO existence}, \textit{Chemtrail conspiracy theory}, \textit{Use of masks in public spaces}, \textit{Sustainable Development Goals}. 

\end{document}